\documentclass[10pt,twocolumn,letterpaper]{article}

\usepackage{cvpr}
\usepackage{xcolor}
\usepackage{times}
\usepackage{epsfig}
\usepackage{graphicx}
\usepackage{amsmath}
\usepackage{amssymb}
\usepackage{adjustbox}
\usepackage{tabularx}
\usepackage{algorithm}
\usepackage{algpseudocode}
\usepackage{multirow}
\usepackage{subcaption}
\usepackage{amsmath,amsfonts,bm}
\graphicspath{{data/}}
\newcolumntype{s}{D{.}{.}{1.2}}
\newcolumntype{d}{D{.}{.}{2.1}}

\newcolumntype{A}{>{\centering\arraybackslash}X}

\graphicspath{{data/}}
\usepackage{amssymb}
\usepackage{pifont}

\newcolumntype{b}{>{\hsize=2.3\hsize}X}
\newcolumntype{s}{>{\hsize=.45\hsize}X}
\newcolumntype{m}{>{\hsize=.9\hsize}X}

\newcommand{\cmark}{\ding{51}}%
\newcommand{\xmark}{\ding{55}}%

\def\mH{{\bm{H}}}

\def\mK{{\bm{K}}}

\def\mQ{{\bm{Q}}}

\def\mV{{\bm{V}}}

\usepackage[pagebackref=true,breaklinks=true,letterpaper=true,colorlinks,bookmarks=false]{hyperref}

\cvprfinalcopy 

\ifcvprfinal\fi
\begin{document}

\title{Multimodal Learning for Hateful Memes Detection}

\author{Yi Zhou$^\ast$, \ \  Zhenhao Chen\thanks{Equal contribution}\\
	$^1$IBM, Singapore  \ \   $^2$The University of Maryland, USA \\
    {\tt\small\textit{joannezhouyi@gmail.com, zhenhao.chen@marylandsmith.umd.edu}}
}

\maketitle
\begin{abstract}
	Memes are used for spreading ideas through social networks. Although most memes are created for humor, some memes become hateful under the combination of pictures and text. Automatically detecting the hateful memes can help reduce their harmful social impact. Unlike the conventional multimodal tasks, where the visual and textual information is semantically aligned, the challenge of hateful memes detection lies in its unique multimodal information. The image and text in memes are weakly aligned or even irrelevant, which requires the model to understand the content and perform reasoning over multiple modalities. In this paper, we focus on multimodal hateful memes detection and propose a novel method that incorporates the image captioning process into the memes detection process. We conduct extensive experiments on multimodal meme datasets and illustrated the effectiveness of our approach. Our model achieves promising results on the Hateful Memes Detection Challenge\footnote{Code is available at \href{https://github.com/czh4/hateful-memes}{GitHub}}.
\end{abstract}
\section{Introduction}\label{sec:intro}
Automatically filter hateful memes is crucial for social networks since memes are spreading on social media, and hateful memes bring attach to people. Given an image, the multimodal meme detection task is expected to find clues from the sentences in the meme image and associate them to the relevant image regions to reach the final detection. Due to the rich and complicated mixture of visual and textual knowledge in memes, it is hard to identify the implicit knowledge behind the multimodal memes efficiently. Driven by the recent advances in neural networks, there have been some works try to detect offensive or misleading content for visual and textual content~\cite{Fortuna2018}. However, current methods are still far from mature because of the huge gap between the meme image and text content.

\begin{figure}[t!]
	\centering     
	\includegraphics[width=\linewidth]{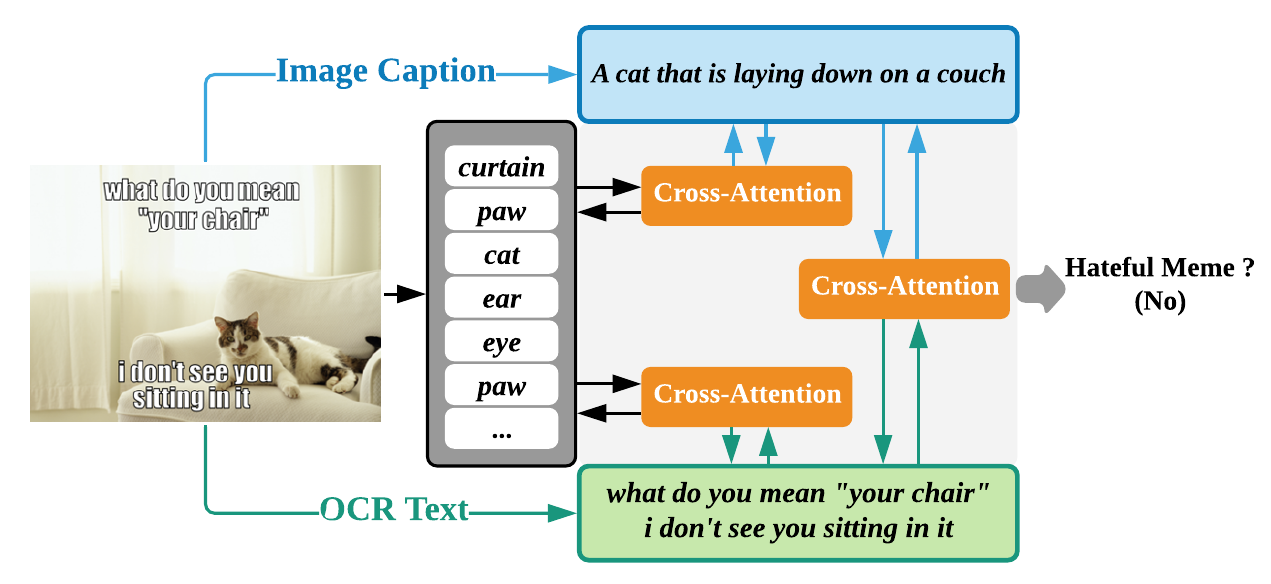}
	\caption{Illustration of our proposed multimodal memes detection approach. It consists of an image captioner, an object detector, a triplet-relation network, and a classifier. Our method considers three different knowledge extracted from each meme: image caption, OCR sentences, and visual features. The proposed triplet-relation network models the triplet-relationships among caption, objects, and OCR sentences, adopting the cross-attention model to learn the more discriminative features from cross-modal embeddings.}
	\label{fig:top}
	\vspace{-5mm}
\end{figure}

Hateful memes detection can be treated as a vision-language (VL) task.  Vision and language problems have gained a lot of attraction in recent years\cite{gu2018recent,gu2020self}, with significant progress on important problems such as Visual Question Answering (VQA)~\cite{Antol2015vqa,Goyal2017vqa2} and Image Captioning~\cite{Chen2015coco,gu2019unpaired,gu2018unpaired,gao2020unsupervised,gu2017empirical}. Specifically, the multimodal memes detection task shares the same spirit as VQA, which predicts the answer based on the input image and question. Recent VQA has been boosted by the advances of image understanding and natural language processing (NLP)~\cite{anderson2017bottomup, peng2018dynamic}. Most recent VQA methods follow the multimodal fusion framework that encodes the image and sentence and then fuses them for answer prediction. Usually, the given image is encoded with a Convolutional Neural Network (CNN) based encoder, and the sentence is encoded with a Recurrent Neural Network (RNN) based encoder. With the advancement of Transformer \cite{vaswani2017attention} network, many recent works incorporate multi-head self-attention mechanisms into their methods~\cite{lu2019vilbert, tan2019lxmert}, and achieve a considerable jump in performances. The core part of the transformer lies in the self-attention mechanism, which transforms input features into contextualized representations with multi-head attention, making it an excellent framework to share information between different modalities.

While a lot of multimodal learning works focus on the fusion between high-level visual and language features~\cite{kim2016hadamard, kim2018bilinear, benyounes2019block, Yu_2018}, it is hard to apply the multimodal fusion method to memes detection directly since memes detection is more focused on the reasoning between visual and textual modalities. Modeling the relationships between multiple modalities and explore the implicit meaning behind them is not an easy task. For example, Fig.~\ref{fig:top}, shows a cat lying down on a couch, but the sentences in the image are not correlated to the picture. The misaligned semantic information between visual and textual features brings significant challenges for memes detection. Even for a human, such an example is hard to identify. Besides, merely extracting the visual and textual information from an image is crucial for memes detection. 

Considering the big gap between visual content and sentence in the meme image, in this paper, we propose a novel approach that generates relevant image descriptions for memes detection. Specifically, we adopt a triplet-relation network to model the relationships between three different inputs. To summarize, our contributions to this paper are twofold. First, we design a Triplet-Relation Network (TRN) that enhances the multimodal relationship modeling between visual regions and sentences. Second, we conduct extensive experiments on the meme detection dataset, which requires highly complex reasoning between image content and sentences.  Experimental evaluation of our approach shows significant improvements in memes detection over the baselines. Our best model also ranks high in the hateful memes detection challenge leaderboard.

\section{Related Works}
\noindent\textbf{Hate Speech Detection.}
Hate speech is a broadly studied problem in network science \cite{Ribeiro2018hatefulusers} and NLP~\cite{Waseem2017understanding,Schmidt2017survey}.
Detecting hate information in language has been studied for a long time. One of the main focuses of hate speech with very diverse targets has appeared in social networks~\cite{hate_mesure,hate_twitter2}.
The common method for hate speech detection is to get a sentence embedding and then feed the embedding into a binary classifier to predict hate speech~\cite{Kumar2018benchmarking,Malmasi2018challenges}.
Several language-based hate speech detection datasets have been released~\cite{Waseem2016racist,Waseem2016hate,Founta2018abusive}. However, the task of hate speech detection has shown to be challenging, and subject to undesired bias \cite{Sap2019acl,Davidson2019acl}, notably the definition of hate speech~\cite{Waseem2017understanding}, which brings the challenges for the machine to understand and detect them. Another direction of hate speech detection targets multimodal speech detection. In~\cite{Hosseinmardi2015cyberbull}, they collected a multimodal dataset based on Instagram images and associated comments. In their dataset, the target labels are created by asking the Crowdflower workers the questions. Similarly, in \cite{Zhong2016cyberbull}, they also collected a similar dataset from Instagram. In \cite{Yang2019exploring}, they found that augment the sentence with visual features can hugely increase the performance of hate speed prediction. Recently, there are some larger dataset proposed. Gomez \etal \cite{Gomez2020exploring} introduced a large dataset for multimodal hate speech detection based on Twitter data. The recent dataset introduced in~\cite{kiela2020hateful} is a larger dataset, which is explicitly designed for multimodal hate speech detection.

\vspace{2mm}
\noindent\textbf{Visual Question Answering.}
Recently, many VQA datasets have been proposed~\cite{ren2015exploring,krishna2017visual}. Similar to multimodal meme detection, the target of VQA is to answer a question given an image. Most current approaches focus on learning the joint embedding between the images and questions~\cite{fukui2016multimodal,yu2017multi,kim2018bilinear}. More specifically, the image and question are passed independently through the image encoder and sentence encoder. The extracted image and question features can then be fused to predict the answer. Usually, those questions are related to the given images, then the challenge of VQA lies in how to reason over the image based on the question.  Attention is one of the crucial improvements to VQA~\cite{zhu2016visual7w,xu2016ask}. In~\cite{xu2016ask}, they first introduce soft and hard attention mechanism. It models the interactions between image regions and words according to their semantic meaning. In~\cite{yang2016stacked}, they propose a stack attention model which increasingly concentrates on the most relevant image regions. In~\cite{lu2016hierarchical}, they present a co-attention method, which calculates attention weights on both image regions and words. Inspired by the huge success of transformer~\cite{vaswani2017transformer}  in neural machine translation (NMT), some works have been proposed to use the transformer to learn cross-modality encoder representations\cite{lu2019vilbert,tan2019lxmert,li2019visualbert}.

Despite the considerable success of vision-language pre-training in VQA. Memes detection is still hard to solve due to its special characters. Apply the methods in VQA to memes detection will facing some issues. First, the sentences in a meme are not like questions, which are mostly based on the image content. Second, it is hard to predict the results from visual modality or textual modality directly. The model needs to understand the implicit relationships between image contents and sentences. Our work adopts image captioning as the bridge which connects the visual information and textual information and models their relationships with a novel triplet-relation network.

\begin{figure*}[htp!]
	\centering
	\includegraphics[width=\linewidth]{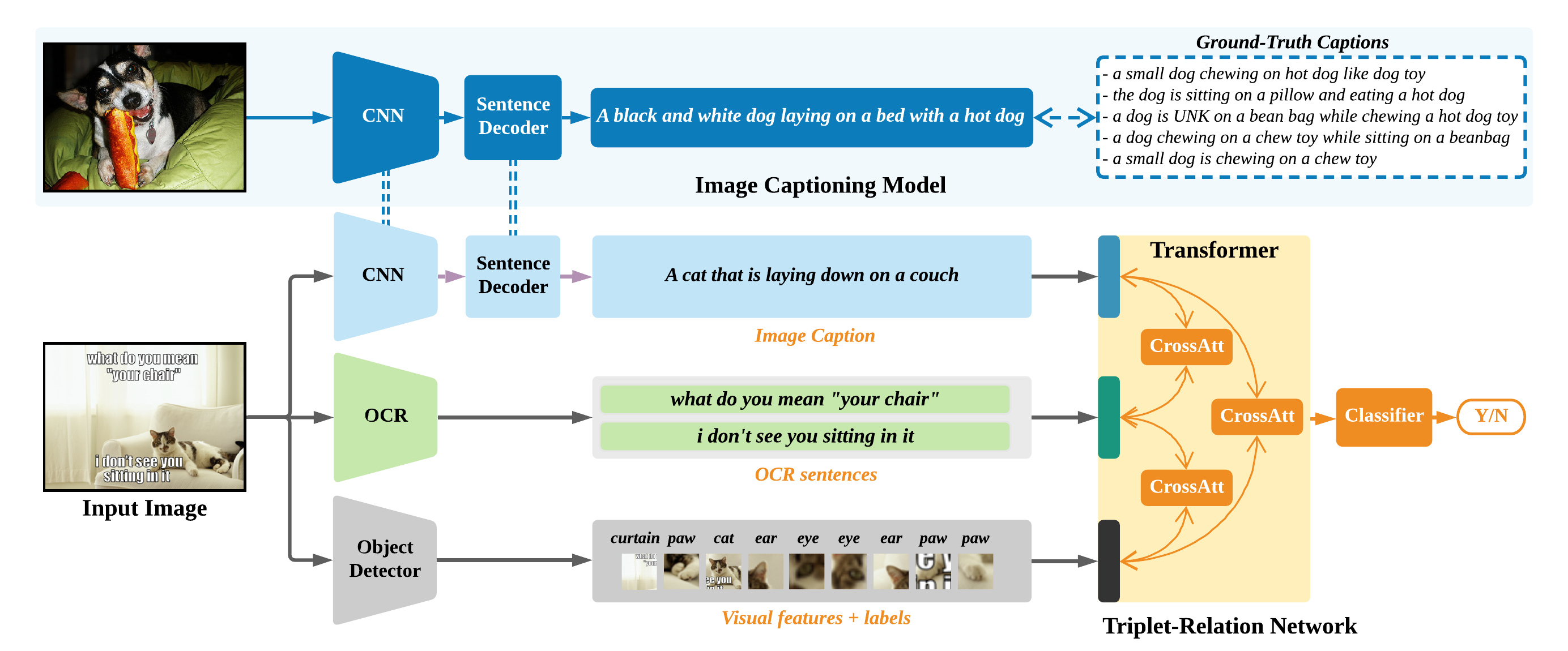}
	\vspace{-5mm}
	\caption{Overview of our proposed hateful memes detection framework.  It consists of three components:  image captioner, object detector, and triplet-relation network. The top branch shows the training of the image captioning model on image-caption pairs. The bottom part is meme detection. It takes image caption, OCR sentences, and object detection results inputs and uses the three inputs' joint representation to predict the answer.}
	\label{fig:overall}
	\vspace{-5mm}
\end{figure*}

\section{Method}

Fig.~\ref{fig:overall} shows our proposed framework. The whole system consists of two training paths: image captioning and multimodal learning. The first path (top part) is identical to the image captioning that maps the image to sentences. The second training path (the bottom part) detects the label from the generated caption, OCR sentences, and detected objects. In the following, we describe our framework in detail.

\subsection{Input Embeddings}
\subsubsection{Sentence Embedding}
The motivation we want to generate an image caption for each meme is that image captioning provides a good solution for image content understanding. The goal for image captioning training is to generate a sentence $S^c$ that describes the content of the image $I$. In particular, we first extract the image feature $\mathbf{f}_I$ with an image encoder $P(\mathbf{f}_I|I)$, and then decode the visual feature into a sentence with a sentence decoder $P(S|\mathbf{f}_I)$. More formally, the image captioning model $P(S|I)$ can be formulated as $P(\mathbf{f}_I|I)P(S|\mathbf{f}_I)$. During inference, the decoding process can be formulated as:
\begin{equation}\label{eq:i2t}
\hat{S}^c = \arg\max_S P(S|\mathbf{f}_v)P(\mathbf{f}_v|I)
\end{equation}
where $\hat{S}^c$ is the predicted image description. The most common loss function to train Eq.~\ref{eq:i2t} is to minimize the negative probability of the target caption words.

As shown in Fig.~\ref{fig:top}, our model has two kinds of textual inputs: image caption $S^c$ and OCR sentence $S^o$.
The predicted caption $\hat{S}^c$ is first split into words $\{\hat{w}^c_1, \ldots, \hat{w}^c_{N_C}\}$ by WordPiece tokenizer~\cite{wu2016google}, where $N_C$ is the number of words. Following the recent vision-language pretraining models~\cite{li2019visualbert,lu2019vilbert}, the textual feature is composed of word embedding, segment embedding, and position embedding:
\begin{equation}
\mathbf{\hat{w}}^c_i=\text{LN}\big(f_{\text{WordEmb}}(\hat{w}^c_i)+f_{\text{SegEmb}}(\hat{w}^c_i)+f_{\text{PosEmb}}(i)\big)\label{eq:emb1}
\end{equation}
where $\mathbf{\hat{w}}^c_i\in \mathbb{R}^{d_w}$ is the word-level feature, $\text{LN}$ represent the layer normalization~\cite{ba2016layer}, $f_{\text{WordEmb}}(\cdot)$, $f_{\text{SegEmb}}(\cdot)$, and $f_{\text{PosEmb}}(\cdot)$ are the embedding functions.

Each meme also contains textual information. We can extract the sentences with the help of the off-the-shelf OCR system. Formally, we can extract the $S^o=\{w^o_1,\ldots, w^o_{N_O}\}$ from the given meme image, where $N_o$ is the number of words. We follow the same operations as image caption and calculate the feature for each token as:
\begin{equation}
\mathbf{{w}}^o_i=\text{LN}\big(f_{\text{WordEmb}}({w}^o_i)+f_{\text{SegEmb}}({w}^o_i)+f_{\text{PosEmb}}(i)\big)\label{eq:emb2}
\end{equation}
where $\mathbf{\hat{w}}^o_i\in \mathbb{R}^{d_w}$ is the word-level feature for OCR token. Those three embedding functions are shared with Eq.~\ref{eq:emb1}.

Following the method in~\cite{li2019visualbert}, we insert the special token [SEP] between and after the sentence. In our design, we concatenate the embeddings of the image caption along with OCR sentence as $\{\mathbf{w}^o_{1:N_O}, \mathbf{w}_{\text{[SEP]}}, \mathbf{\hat{w}}^c_{1:N_C}, \mathbf{w}_{\text{[SEP]}}\}$, where $\mathbf{w}_{\text{[SEP]}}$ is the word-level embedding for special token.

\subsubsection{Image Embedding}
Instead of getting the global representation for each image, we take the visual features of detected objects as the representation for the image. Specifically, we extract and keep a fixed number of semantic region proposals from the pretrained Faster R-CNN\cite{ren2015faster}\footnote{https://github.com/airsplay/py-bottom-up-attention}.
Formally, an image $I$ consists of $N_v$ objects, where each object $o_i$ is represented by its region-of-interest (RoI) feature $\mathbf{v}_i\in \mathbb{R}^{d_o}$, and its positional feature $\mathbf{p}_i^o\in \mathbb{R}^4$ (normalized top-left and bottom-right coordinates). Each region embedding is calculated as follows:
\begin{equation} 
\mathbf{v}^o_i = \text{LN}\left(f_{\text{VisualEmb}}(\mathbf{v}_i)+f_{\text{VisualPos}}( \mathbf{p}_i^o)\right)\label{eq:viusal}
\end{equation}
where $\mathbf{v}^o_i\in \mathbb{R}^{d_v}$ is the position-aware feature for each proposal, $f_{\text{VisualEmb}}(\cdot)$ and $f_{\text{VisualPos}}(\cdot)$ are two embedding layers.

\subsection{Triplet-Relation Network}
The target of triplet-relation network is to model the cross-modality relationships between image features ($\mathbf{v}^o_{1:N_v}$) and two textual features ($\mathbf{\hat{w}}^c_{1:N_c}$ and $\mathbf{w}^o_{1:N_o}$).
Motivated by the success of the self-attention mechanism~\cite{vaswani2017attention}, we adopt the transformer network as the core module for our TRN.

Each transformer block consists of three main components: Query ($\mQ$), Keys ($\mK$), and Values ($\mV$).
Specifically, let $\mH^l = \{h_1, \ldots, h_N\}$ be the encoded features at $l$-th layer.
$\mH^l$ is first linearly transformed into $\mQ^l$, $\mK^l$, and $\mV^l$ with learnable parameters. The output $\mH^{l+1}$ is calculated with a softmax function to generate the weighted-average score over its input values. For each transformer layer, we calculate the outputs for each head as follows:
\begin{equation}\label{eq:softmax}
\mH^{l+1}_{\text{Self-Att}} = \text{Softmax}(\frac{\mQ^{l}({\mK}^{l})^T}{\sqrt{d_k}}) \cdot \mV^{l}
\end{equation}
where $d_k$ is the dimension of the Keys and $\mH^{l+1}_{\text{Self-Att}}$ is the output representation for each head.

Note that the inputs $\mH^0$ to the TRN are the combination of the two textual features and visual features.
In this paper, we explore two variants of TRN: one-stream~\cite{li2019visualbert} and two-stream~\cite{lu2019vilbert}. One-stream denotes that we model the visual and textual features together in a single stream. Two-stream means we use two separate streams for vision and language processing that interact through co-attentional transformer layers. For each variant, we stack $L_{\text{TRN}}$ these attention layers which serve the function of discovering relationships from one modality to another. For meme detection, we take the final representation $h_{\text{[CLS]}}$ for the $\text{[CLS]}$ token as the joint representation.

\subsection{Learning}

\noindent\textbf{Training image captioner.}
For the image captioner learning, the target is to generate image descriptions close to the ground-truth captions. Follow the recent image captioning methods; we first train image encoder and sentence decoder by minimizing the cross-entropy (XE) loss as follows:
\begin{equation}
\mathcal{L}_{\text{XE}}  = - \sum_i \log p(w^c_i|w^c_{0:i-1}, I) \label{equ:equ_celoss}
\end{equation}
where $ p(w^c_t|w^c_{0:t-1})$ is the output probability of $t$-th word in the sentence given by the sentence decoder.

After training the image captioner with Eq.~\ref{equ:equ_celoss}, we further apply a reinforcement learning (RL) loss that takes the CIDEr~\cite{vedantam2015cider} score as a reward and optimize the image captioner by minimizing the negative expected rewards as:
\begin{equation}
\mathcal{L}_{\text{RL}}  = -\mathbb{E}_{\tilde{S}^c \sim P_{\theta}}[r(\tilde{S}^c)]
\label{equ:raw_rl}
\end{equation}
where $\tilde{S}^c=\{\tilde{w}^c_{1:N_c}\}$ is the sampled caption, $r(\tilde{S}^c)$ is the reward function calculated by the CIDEr metric, $\theta$ is the parameter for image captioner. Following the SCST method described in~\cite{rennie2016self}, Eq.~\ref{equ:raw_rl} can be approximated as:
\begin{equation}
\nabla_{\theta}\mathcal{L}_{\text{RL}}(\theta)\approx-(r(\tilde{S}^c) - r(\hat{S}^c))\cdot \nabla_{\theta}\log p(\tilde{S}^c)\label{eq:rl_loss_one_stage}
\end{equation}
where $\hat{S}^c$ is the image caption predicted by greedy decoding. The relative reward design in Eq.~\ref{eq:rl_loss_one_stage} tends to increase the probability of $\tilde{S}^c$ that score higher than $\hat{S}^c$.

\vspace{2mm}
\noindent\textbf{Training classifier.}
For meme detection training, we feed the joint representation $h_{\text{[CLS]}}$ of language and visual content to a fully-connected (FC) layer, followed by a softmax layer, to get the prediction probability $\hat{y}=\text{softmax}(f_{\text{FC}}(h_{\text{[CLS]}}))$. 
A binary cross-entropy (BCE) loss function is used as the final loss function for meme detection:
\begin{equation}
\mathcal{L_{\text{BCE}}} =  -\mathbb{E}_{I\sim \mathcal{D}}[ y \log (\hat{y}) +  (1-y) \log(1 - \hat{y})]
\end{equation}
where $N$ is the number of training samples, $I$ is sampled from the training set $\mathcal{D}$, $y$ and  $\hat{y}$ represent the ground-truth label and detected result for the meme, respectively.

\begin{table*}[ht!]
	\small
	\caption{Experimental results and comparison on Hateful Memes Detection dev and test splits (\textbf{bold} numbers are the best results). `OCR Text (Back-Translation)' means we augment the OCR sentences in training set through trained back-translators. `Image Caption' means using captions for meme detection.  AUROC in percentage (\%) are reported.}
	\vspace{-3mm}
	\begin{center}
		\setlength{\tabcolsep}{6pt}
		\begin{tabular}{c|c|c|c|c|c}
			\hline
			\multirow{2}{*}{\textbf{Model}}& \multicolumn{1}{c|}{\textbf{Basic Inputs}}&\multicolumn{3}{c|}{\textbf{Additional Inputs}}&  \multirow{2}{*}{\textbf{AUROC (Dev)}}\\
			\cline{2-5}
			& \textbf{Object Feature + OCR Text}& \textbf{Object Labels}&\textbf{OCR Text (Back-Translation)}& \textbf{Image Caption}&    \\
			\hline
			\multirow{8}{*}{\textbf{V+L}}
			& \cmark & \xmark & \xmark  & \xmark & 70.47  \\
			& \cmark & \xmark & \xmark  & \cmark & 72.97  \\\cline{2-6}
			
			& \cmark & \xmark & \cmark  & \xmark & 72.43 \\
			& \cmark & \xmark & \cmark  & \cmark & \textbf{73.93}  \\\cline{2-6}
			
			& \cmark & \cmark & \xmark  & \xmark & 70.96  \\
			& \cmark & \cmark & \xmark  & \cmark & 72.66  \\\cline{2-6}
			
			& \cmark & \cmark & \cmark  & \xmark & 71.57  \\
			& \cmark & \cmark & \cmark  & \cmark & 72.15  \\
			\hline
			\hline
			\multirow{8}{*}{\textbf{V\&L}}
			& \cmark & \xmark & \xmark  & \xmark & 66.94  \\
			& \cmark & \xmark & \xmark  & \cmark & \textbf{71.11} \\\cline{2-6}
			
			& \cmark & \xmark & \cmark  & \xmark & 63.47 \\
			& \cmark & \xmark & \cmark  & \cmark & 67.94 \\\cline{2-6}
			
			& \cmark & \cmark & \xmark  & \xmark & 70.22 \\
			& \cmark & \cmark & \xmark  & \cmark & 70.46 \\\cline{2-6}
			
			& \cmark & \cmark & \cmark  & \xmark & 66.68 \\
			& \cmark & \cmark & \cmark  & \cmark & 69.85 \\
			\hline
		\end{tabular}
	\end{center}\label{tab:baseline}
	\vspace{-8mm}
\end{table*}

\section{Experiments}
\subsection{Dataset and Preprocessing}
In our experiments, we use two datasets: MSCOCO~\cite{lin2014microsoft} and Hateful Memes Detection Challenge dataset provided by Facebook~\cite{kiela2020hateful}. We describe the detail of each dataset below.

\vspace{2mm}
\noindent\textbf{MSCOCO.} MSCOCO is an image captioning dataset which has been widely used in image captioning task~\cite{Xu2015show,gu2019graph}. It contains 123,000 images, where each image has five reference captions. During training, we follow the setting of \cite{karpathy2015deep} by using  113,287 images for training, 5,000 images for validation, and 5,000 images for testing. The best image captioner is selected base on the highest CIDEr score.

\vspace{2mm}
\noindent\textbf{The Hateful Memes Challenge Dataset.} This dataset is collected by Facebook AI as the challenge set. The dataset includes 10,000 memes, each sample containing an image and extracted OCR sentence in the image. For the purpose of this challenge, the labels of memes have two types, non-hateful and hateful. The dev and test set consist of 5\% and 10\% of the data respectively and are fully balanced, while the rest train set has 64\% non-hateful memes and 36\% hateful memes.

\vspace{2mm}
\noindent\textbf{Data Augmentation.}
We augment the sentence in the hateful memes dataset with the back-translation strategy. Specifically, we enrich the OCR sentences through two pretrained back-translator\footnote{https://github.com/pytorch/fairseq}: English-German-English and English-Russian-English. We also apply different beam sizes (2, 5, and 10) during the sentence decoding to get the diverse sentences for each sentence.

\subsection{Implementation Details}
We present the hyperparameters related to our baselines and discuss those related to model training. For visual feature preprocessing, we extract the ROI features using the pretrained Faster R-CNN object detector~\cite{ren2015faster}. We keep a fixed number of region proposals for each image and get the corresponding RoI feature, bounding box, and predicted labels.
During image captioning training, we use a mini-batch size of 100. The initial learning rate is 1e-4 and gradually increased to 4e-5. We use Adam \cite{kingma2014adam} as the optimizer. We use the pretrained ResNet101~\cite{he2015deep} as the image encoder and randomly initialize the weights for the sentence decoder.
During memes detection training, we set the dimension of the hidden state of the transformer to 768, and initialize the transformer in our model with the BERT models pertained in MMF\footnote{Https://github.com/facebookresearch/mmf}.
We use Adam~\cite{kingma2014adam} optimizer with an initial learning rate of 5e-5 and train for 10,000 steps.
We report the performance of our models with the Area Under the Curve of the Receiver Operating Characteristic (AUROC). It measures how well the memes predictor discriminates between the classes as its decision threshold is varied. During online submission, we submit predicted probabilities for the test samples. The online (Phase1 and Phase2) rankings are decided based on the best submissions by AUROC.

\begin{table}[ht]
	\centering\small
	\caption{Performance comparison with existing methods on online test server.}
	\vspace{-3mm}
	\setlength{\tabcolsep}{2pt}{
		\begin{tabular}{c|l|cc}
			\hline
			\textbf{Inputs} & \textbf{Model} & \textbf{AUROC (Test)}\\\hline
			\multirow{3}{*}{Image} & Human~\cite{kiela2020hateful} & 82.65\\\hline
			& Image-Grid~\cite{kiela2020hateful} & 52.63\\
			& Image-Region~\cite{kiela2020hateful} &55.92\\\hline
			\multirow{1}{*}{Text} & Text BERT~\cite{kiela2020hateful} & 65.08\\\hline
			\multirow{8}{*}{\shortstack{Image + Text}} & Late Fusion & 64.75\\
			& Concat BERT~\cite{kiela2020hateful} & 65.79\\
			& MMBT-Grid~\cite{kiela2020hateful} & 67.92\\
			& MMBT-Region~\cite{kiela2020hateful} & 70.73\\
			& ViLBERT~\cite{kiela2020hateful} & 70.45\\
			& Visual BERT~\cite{kiela2020hateful} & 71.33\\
			& ViLBERT CC~\cite{kiela2020hateful} & 70.03\\
			& Visual BERT COCO~\cite{kiela2020hateful} & 71.41\\
			\hline
			\multirow{2}{*}{\shortstack{Image + Text + Caption}} & Ours (V+L) & \textbf{73.42}\\
			& Ours (V\&L) & \textbf{74.80}\\
			\hline
		\end{tabular}
	}
	\label{tab:results}
	\vspace{-5mm}
\end{table}

\subsection{Result and Discussion}
We conduct the ablation study in Table~\ref{tab:baseline}.
The baseline models can be divided into two categories: V+L and V\&L. V+L represents the one-stream model. It takes the concatenated visual and textual features as input and produces contextually embedded features with a single BERT. The parameters are shared between visual and textual encoding. V\&L represents the two-stream model. It first adopts the two-stream for each modality and then models the cross-relationship with a cross-attention based transformer. In our experiments, we initialize the V+L models with pretrained Visual BERT~\cite{li2019visualbert}, and the V\&L models with pretrained ViLBERT~\cite{lu2019vilbert}. The parameters for all models are finetuned on the meme detection task.

\begin{figure*}[ht!]
	\begin{center}
	\end{center}
	\includegraphics[width=1.0\linewidth]{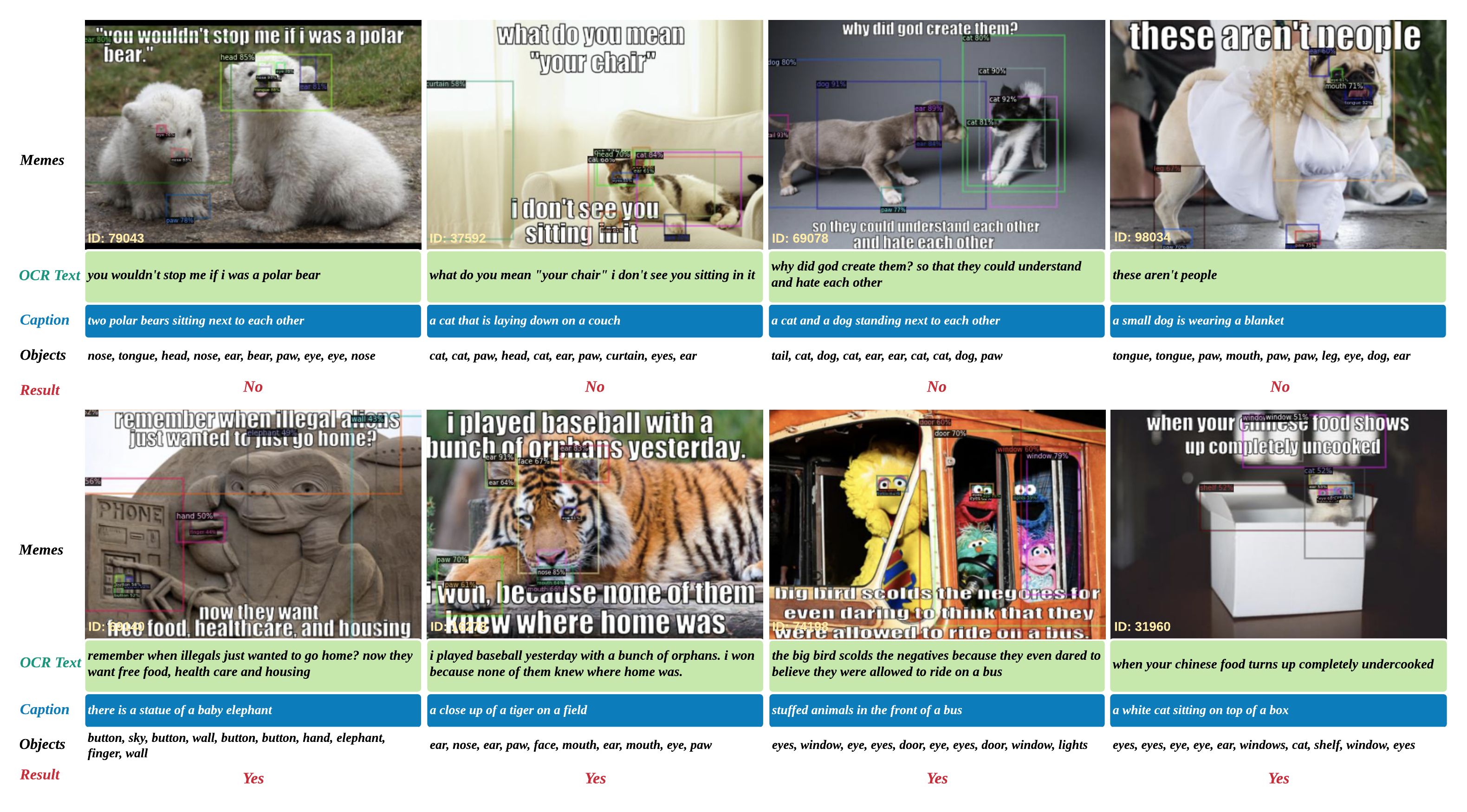}
	\centering
	\vspace{-8mm}
	\caption{Qualitative examples of of hateful memes detection. We show the generated caption and object labels for each sample.}
	\vspace{-5mm}
	\label{fig:visualizations}
\end{figure*}

\vspace{2mm}
\noindent\textbf{Effectiveness of Image Captioning.}
In Table~\ref{tab:baseline}, we can see that V\&L models with image caption outperform other V\&L models by a large margin on the dev set. These results support our motivation that image captioning helps the hateful memes detection. The generated image descriptions provide more meaningful clues to detect the `hateful' memes since the captions can describe the image's content and provide semantic information for cross-modality relationship reasoning. The performance boost brings by image captioning further indicates that, due to the rich and societal content in memes,  only considering object detection and OCR recognition is insufficient. A practical solution should also explore some additional information related to the meme.

\vspace{2mm}
\noindent\textbf{Effectiveness of Language Augmentation.}
We also verify the effectiveness of data augmentation in Table~\ref{tab:baseline}. We can see that back-translation can bring some improvement to V\&L models. However, for V+L models, back-translation does not show improvement. We think the reason for the ineffectiveness of back-translation on V+L models is that the one-stream models handle the multimodal embeddings with the shared BERT model. For hateful memes detection, the OCR sentences are not semantically aligned with the image content, which weakens sentence augmentation effectiveness.  For V\&L, introducing augmented sentences can improve the intra-modality modeling for language since it contains independent branches for visual and textual modeling separately.

\vspace{2mm}
\noindent\textbf{Effectiveness of Visual Labels.}
We consider combining the predicted object labels as additional input features. Specifically, we treat the object labels as linguistic words and concatenate them with OCR sentence and image caption. We can see that the object labels can improve the V+L and V\&L models. This is reasonable since object labels can be seen as the ``\textit{anchor}" between RoI features and textual features (OCR text and caption). A similar finding can also be found in~\cite{li2020oscar}.

\vspace{2mm}
\noindent\textbf{Comparisons with the Existing Methods.}
Table~\ref{tab:results} shows the comparisons of our method on Hateful memes challenge with existing methods. We can see that our method achieves better performance, which demonstrates the advantage of our triplet-relation network. We also submit the results of our model to the online testing server, and our best model with ensembling (named as ``\textit{naoki}") achieves the 13th position among 276 teams in the Phase 2 competition\footnote{https://www.drivendata.org/competitions/70/hateful-memes-phase-2/leaderboard/}.

\vspace{2mm}
\noindent\textbf{Visualization Results.}
Fig.~\ref{fig:visualizations} shows some generated captions and predicted results. From those samples, we can see that our model can understand the implicit meaning behind the image and sentences with the help of the predicted image captions. For example, in the last image, although there is no explicit relationship between the image and OCR test, our method still can predict the correct result by connecting different modalities with the image caption.

\section{Conclusion}
In this paper, we propose a novel multimodal learning method for hateful memes detection. Our proposed model exploits the combination of image captions and memes to enhance cross-modality relationship modeling for hateful memes detection. We envision such a triplet-relation network to be extended to other frameworks that require additional features from multimodal signals. Our model achieves competitive results in the hateful memes detection challenge.

\bibliographystyle{IEEEbib}
\bibliography{ref}

\end{document}